# Total Variation Applications in Computer Vision

Vania V. Estrela[*1] , Hermes A. Magalhães [2], Osamu Saotome [3]

[1]Departamento de Telecomunicacoes, Universidade Federal Fluminense (UFF), Niterói, RJ, Brazil, vania.estrela.phd@ieee.org

[2]Universidade Federal de Minas Gerais, Escola de Engenharia, Departamento de Engenharia Eletrônica. Av Antônio Carlos, 6627, DELT-EEUFMG, Sala 2506, Belo Horizonte, MG, Brazil, hermes@cefala.org

[3]ITA - Instituto Tecnólogico de Aeronáutica, CTA-ITA-IEEA 12.228-900 , São José dos Campos,SP, Brazil, osaotome@gmail.com

## ABSTRACT

*The objectives of this chapter are: (i) to introduce a concise overview of regularization; (ii) to define and to explain the role of a particular type of regularization called total variation norm (TV-norm) in computer vision tasks; (iii) to set up a brief discussion on the mathematical background of TV methods; and (iv) to establish a relationship between models and a few existing methods to solve problems cast as TV-norm. For the most part, image-processing algorithms blur the edges of the estimated images, however TV regularization preserves the edges with no prior information on the observed and the original images. The regularization scalar parameter λ controls the amount of regularization allowed and it is an essential to obtain a high-quality regularized output. A wide-ranging review of several ways to put into practice TV regularization as well as its advantages and limitations are discussed.*

Keywords: Total Variation, Regularization, Computer Vision, Machine Learning, Variational Methods, Computational Intelligence, Image Processing.

## 1. INTRODUCTION

This chapter investigates robustness properties of machine learning (ML) methods based on convex risk minimization applied to computer vision. Kernel regression, support vector machines (SVMs), and least squares (LS) can be regarded as special cases of ML. The minimization of a regularized empirical risk based on convex functionals has an essential role in statistical learning theory (Vapnik, 1995), because (i) such classifiers are generally consistent under weak conditions; and (ii) robust statistics investigate the impact of data deviations on the results of estimation, testing or prediction methods.

In practice, one has to apply ML methods - which are nonparametric tools - to a data set with a finite sample size. Even so, the robustness issue is important, because the assumption that all data points were independently generated by the same distribution can be contravened and outliers habitually occur in real data sets.

The real use of regularized learning methods depends significantly on the option to put together intelligent models fast and successfully, besides calling for efficient optimization methods. Many ML algorithms involve the ability to compare two objects by means of the similarity or distance between them. In many cases, existing distance or similarity functions such as the Euclidean distance are enough. However, some problems require more appropriate metrics. For instance, since the Euclidean distance uses of the $L_2$-norm, it is likely to perform scantily in the presence of outliers. The Mahalanobis distance is a straightforward and all-purpose method that subjects data to a linear transformation. Notwithstanding, Mahalanobis distances have two key problems: 1) the parameter vector to be learned increases quadratically as data grows, which poses a problem related to dimensionality; and 2) learning a linear transformation is not sufficient for data sets with nonlinear decision boundaries.

Models can also be selected by means of regularization methods, that is, they are penalizing depending on the number of parameters (Alpaydin, 2004) (Fromont, 2007). Generally, Bayesian learning techniques make use of knowledge on the prior probability distributions in order to assign lower probabilities to models that are more complicated. Some popular model selection techniques are the Akaike information criterion (AIC), the Takeuchi information criterion (TIC), the Bayesian information criterion (BIC), the cross-validation technique (CV), and the minimum description length (MDL).

This chapter aims at showing how Total Variation (TV) regularization can be practically implemented in order to solve several computer vision applications although is still a subject under research. Initially, TV has been introduced in (Rudin, Osher, & Fatemi, 1992) and, since then, it has found several applications in computer vision such as image restoration (Rudin & Osher, 1994), image denoising (Matteos, Molina & Katsaggelos, 2005)(Molina, Vega & Katsaggelos, 2007), blind deconvolution (Chan & Wong, 1998), resolution enhancement (Guichard & Malgouyres, 1998), compression (Alter, Durand, & Froment, 2005), motion estimation (Drulea & Nedevschi, 2011), texture segmentation/discrimination (Roudenko, 2004). These applications involve the use of TV regularization that allows selecting the best solution from a set of several possible ones.

## 2. BACKGROUND

## 2.1 Regularization

In machine learning (ML) and inverse problems, regularization brings in extra information to solve an ill-posed problem and/or to circumvent overfitting. Representative information is taken into consideration via insertion of a penalty function based on constraints for solution smoothness or bounds on the vector space norm. Representative cases of regularization in statistical ML involve methods like ridge regression, lasso, and $L_2$-norm for example.

If a problem has (a) a unique solution, and (b) the solution is robust to small data perturbations, then it is called well-posed. When at least one of them is violated, it is named ill-posed and it requires special care.

Non-uniqueness is a consequence of not having enough data on the original model and it is not detrimental at all times. Depending on the desired characteristics of a good solution or some measures of goodness, then an estimate can be picked up from a set of multiple solutions. Nevertheless, if one does not know how to evaluate an estimate, then a very good way to handle non-uniqueness is to enforce some prior information about domain in order to constrain the solution set.

Instability results from an effort to undo cause-effect relations. Solving a forward problem is the most natural way of finding a solution, since cause always goes before effect. In reality, one has access to corrupted measures, which means one aims at finding the cause without a closed-form description of the system being analyzed (system model).

Regularization can be isotropic or anisotropic on the smoothness terms. Isotropic regularization schemes relax smoothness constraints at boundaries. Anisotropic formulations let smoothing occurs along the borders but not transversal to them.

The concept of regularization relies on the use of norms. This chapter will only consider expressions of this form

$$\| x \|_P = \left( \sum_i |x_i|^p \right)^{\frac{1}{p}},$$

(1)

where the most popular ones are described as follows:

$L_2$ norm: $\|x\|_2 = \sqrt{\sum_i x_i^2}$ is also known as Euclidean distance. Algorithms relying on it generate smooth results, which penalize image edges.

$L_1$ (Manhattan) norm: $\|x\|_1 = \sum_i |x_i|$ is the sum of the absolute values of the distances in the original space. Algorithms using this norm preserve image edges, although they are time-consuming.

## 2.2 Least Squares (LS) Regularization

Given a system $\mathbf{g}=\mathbf{Hf}$, where $\mathbf{H}$ is a forward operator, the simplest form of regularizing an ill-posed problem is the linear Least Squares (LS) with Euclidean ($L_2$) norm that aims at minimizing the residual

$\varepsilon^2_{LS} = J(\mathbf{f}) = \| \mathbf{Hf}\text{-}\mathbf{g} \|_2^2$. That is,

(2)

$\mathbf{f_{LS}} = (\mathbf{H^TH})^{-1}\mathbf{H^Tf}$,

(3)

where $J(\mathbf{f})$ is a functional to be minimized, $\varepsilon^2_{LS}$ is the squared of the residuals, $\mathbf{f}$ is an estimate of $\mathbf{f}$ according to the LS squares criterion and $\| . \|_2^2 = \| . \|^2$. If $\mathbf{H}$ is ill-conditioned or singular, then (3) may not be a good estimate. A regularization term $\| \mathbf{Qf} \|^2$ (also known as regularization penalty) is included in this minimization functional, then it will lead to the Regularized Least Squares (RLS) estimate (Coelho & Estrela, 2012)(Coelho & Estrela, 2012)(Kang & Katsaggelos, 1995)(Molina, Vega, & Katsaggelos, 2007). Hence, the new functional becomes

$J(\mathbf{f}) = \varepsilon^2_{RLS} = \| \mathbf{Hf}\text{-}\mathbf{g} \|^2 + \| \mathbf{Qf} \|^2$.

(4)

The most common case is $\mathbf{Q}=\lambda\mathbf{I}$, where $\lambda$ is a scalar regularization parameter and $\mathbf{I}$ is the identity matrix, nevertheless other types of regularization matrix $\mathbf{Q}$ can be chosen in order to enhance problem conditioning (such as a first or a second order differentiation matrix), therefore making possible a better numerical solution given by:

$\mathbf{f_{RLS}} = (\mathbf{H^TH}+\mathbf{Q^TQ})^{-1}\mathbf{H^Tf}$.

(5)

From a Bayesian point of view, this is equivalent to adding some additional assumptions with in order to obtain a stable estimate. Statistically, $\mathbf{f}$ prior is frequently assumed to be zero-mean Gaussian with independent and identically distributed (iid) components with identical standard deviation $\sigma_f$. The data $\mathbf{g}$ are also erroneous, and iid with zero mean and standard deviation $\sigma_g$. When $\mathbf{Q}=\lambda\mathbf{I}$, $\mathbf{f_{RLS}}$ has $\lambda = \sigma_g / \sigma_f$ according to the previous expression.

**Generalized Tikhonov regularization (GTR)**

A more general functional is

$J(\mathbf{f}) = \| \mathbf{Hf}\text{-}\mathbf{g} \|_P^2 + \| \mathbf{f}\text{-}\mathbf{f_0} \|_Q^2$.

(6)

where $\|\mathbf{f}\|_Q^2 = \mathbf{f}^T\mathbf{Q}\mathbf{f}$ is a weighted norm. A Bayesian analysis shows that $\mathbf{P}$ is the inverse covariance matrix of $\mathbf{g}$, $\mathbf{f_0} = E\{\mathbf{f}\}$, and $\mathbf{Q}$ the inverse covariance matrix of $\mathbf{f}$. The Tikhonov matrix $\mathbf{\Lambda}$ is obtained from $\mathbf{Q} = \mathbf{\Lambda^T}\,\mathbf{\Lambda}$ (the Cholesky factorization), and it can be regarded as a whitening filter.

The resulting estimate is $\mathbf{f_{GTR}} = \mathbf{f_0} + (\mathbf{H^TPH} + \mathbf{Q})^{-1}\mathbf{H^TP}(\mathbf{g} - \mathbf{Hf_0})$. The LS and RLS estimates are special cases of the GTR solution (Blomgren & Chan, 1998)(Chan & Wong, 1998).

Usually, the discretization of integral equations lead to discrete ill-conditioned problems, and Tikhonov regularization can be applied in the original infinite dimensional space. The previous expression can be interpreted as follows: $\mathbf{H}$ is a Hilbert space compact operator, plus $\mathbf{f}$ and $\mathbf{g}$ are elements in the domain and range of $\mathbf{H}$ respectively. This implies that the operator $(\mathbf{H*H} + \mathbf{Q^TQ})$ is a self-adjoint as well as a bounded invertible operator.

## 2.3 Total Variation Regularization

Total Variation (TV) regularization is a deterministic technique that safeguards discontinuities in image processing tasks. For a known kernel $\mathbf{H}$, the true image $\mathbf{f}$ satisfies the relationship $\mathbf{g} \approx \mathbf{Hf}$. The approximation symbol accounts for noise. With the purpose of imposing uniqueness and circumvent distortions, the predicted image $\mathbf{f}$ can be described as the minimizer of

A new functional can be stated as

$$J_{TV}(\mathbf{f}) = \frac{1}{2}\|\mathbf{Hu} - \mathbf{g}\|_2^2 + \lambda TV(\mathbf{f}), \qquad (7)$$

where $\mathbf{r} = (x, y)$ is a pixel location, $\nabla\mathbf{f}$ is the gradient of $\mathbf{f}$, $\Omega$ is the corresponding Hilbert space and $\lambda \geq 0$ is a hyperparameter whose value depends on the amount of noise. The TV norm is defined by

$$TV(\mathbf{f}) = \int_\Omega \sqrt{|\nabla\mathbf{f}|^2}\, dxdy \ . \qquad (8)$$

In the previous expression, $\nabla\mathbf{f}$ is the Laplacian of $\mathbf{f}$, the term $\|\mathbf{Hf} - \mathbf{g}\|^2$ is a fidelity (penalty) term. Recently, various efficient implementation of the TV norm have been proposed. Nevertheless, finding the exact value of $\lambda$ is computationally demanding. Despite some difficulties concerning the discretization of the previous equation because it introduces high frequency artifacts in the estimated solution, it can be proven that they can be avoided by different TV discretization strategies (Sfikas, Nikou, Galatsanos, & Heinrich, 2011). The main advantage of TV regularization is the fact that this variational approach has edge-preserving properties, but textures and fine-scale details are still removed. Given that it is not possible to differentiate TV($\mathbf{f}$) at zero, a small constant $\alpha > 0$ is placed in the preceding expression in this fashion:

$$TV(\mathbf{f}) = \int_\Omega \sqrt{|\nabla\mathbf{f}|^2 + \alpha^2}\, dxdy \qquad (9)$$

The TV regularization term allows selecting amid numerous potential estimates the optimal one. With the intention of enforcing uniqueness and evade serious ringing artifacts, the estimated image $\mathbf{f}$ will be the value of $\mathbf{f}$ that minimizes the following functional:

$$J_{TV}(\mathbf{f}) = \frac{1}{2} \, \| \mathbf{Hu\text{-}g} \|_2^2 + \lambda \int_{\Omega} \, \sqrt{|\nabla \mathbf{f}|^2 + \alpha^2} \, dxdy \ . \tag{10}$$

Knowledge on the image discontinuities accounts for the gradient magnitude $\sqrt{|\nabla \mathbf{f}|^2 + \alpha^2} \, dxdy$. To find $\mathbf{f}$ involves two steps: to define a discrete version of $J_{TV}(\mathbf{f})$ for images and to find an algorithm to minimize the discrete problem. Provided the chosen algorithm converges, the solution will depend only on the discretization selected, which is a modeling concern.

Analogously to what happens in RLS (Galatsanos & Katsaggelos, 1992), $\lambda$ is very important when it comes to controlling the amount of noise allowed in the process. If $\lambda{=}0$, then no denoising is applied and the outcome is equal to the original image. On the other hand, as $\lambda \quad \infty$, the TV term becomes progressively stronger, then the output image becomes more different from the original one that is corresponding to having smaller TV. Consequently, the selection of regularization parameter is vital to attain the adequate amount of noise elimination.

Practically speaking, the gradient can be approximated by means of different norms. The TV norm introduced by (Rudin, Osher, & Fatemi, 1992) is named $TV_{fro}$ in this text. It is isotropic, $L_2$-based, and non-differentiable. If the gradient of $\mathbf{f}$ is $\nabla \mathbf{f} = (D^x_{i,j}, \ D^y_{i,j})$ where

$$D^x_{i,j} \mathbf{f} = \frac{f_{i,j} - f_{i\text{-}1,j}}{\Delta x},$$

$$D^y_{i,j} \mathbf{f} = \frac{f_{i,j} - f_{i,j\text{-}1}}{\Delta y}, \text{ and}$$

$x{=}\ y{=}1$, then (9) can be rewritten as

$$TV_{fro}(\mathbf{f}) = \sum_{i,j} \sqrt{|f_{i+1,j} - f_{i,j}|^2 + |f_{i,j+1} - f_{i,j}|^2 + \alpha^2} \ . \tag{11}$$

Since other choices of discretization are possible for the gradient, then an alternative to $TV_{fro}$ relying on the $L_1$-norm for an $M{\times}N$ image can be obtained by taking into consideration the following relationships and approximations:

$$\begin{aligned} TV_{L_1\text{-}fro}(\mathbf{f}) &= \sum_{i,j} \left\{ \sqrt{|f_{i+1,j} - f_{i,j}|^2} + \sqrt{|f_{i,j+1} - f_{i,j}|^2} + \alpha \right\} \\ &= \sum_{i,j} \left\{ |f_{i+1,j} - f_{i,j}| + |f_{i,j+1} - f_{i,j}| + \alpha \right\} \\ &= MN\alpha + \sum_{i,j} \left\{ |f_{i+1,j} - f_{i,j}| + |f_{i,j+1} - f_{i,j}| \right\}. \end{aligned} \tag{12}$$

$TV_{L1\text{-}fro}$ is easier to minimize, it is also anisotropic and less time-consuming. Because unraveling this denoising problem is far from trivial, modern investigation on compressed sensing algorithms such as (Antonin Chambolle, 2004)(Donoho, 2008) (Friedman, Hastie, & Tibshirani, 2010) (Afonso, Bioucas-Dias, & Figueiredo, 2011) solve variants of the original TV-norm problem.

The modification to the $L_p$-norm has a remarkable effect on the calculation of $\mathbf{f}$. It has been shown by P.C.Hansen that the solution consists of polynomial pieces, and the degree of polynomials is $p{-}1$.

The problem can be better stated by defining a function $\Psi(t) = 2\sqrt{t + \alpha^2}$ and rewriting (9), (10), and (12) as follows:

$$TV(\mathbf{f}) = \frac{1}{2}\sum_{i=1}^{M}\sum_{j=1}^{N}\Psi\left(\left(D_{i,j}^x\mathbf{f}\right)^2 + \left(D_{i,j}^y\mathbf{f}\right)^2 + \alpha^2\right)$$

$$\Rightarrow \quad TV(\mathbf{f}) = \frac{1}{2}\sum_{i=1}^{M}\sum_{j=1}^{N}\Psi\left\{|f_{i+1,j} - f_{i,j}| + |f_{i,j+1} - f_{i,j}| + \alpha^2\right\}. \tag{14}$$

Gradient of Total Variation from (14) is given by

$$\Rightarrow \quad \nabla TV(\mathbf{f}) = \frac{1}{2}\sum_{i=1}^{M}\sum_{j=1}^{N}\Psi'\left\{|f_{i+1,j} - f_{i,j}| + |f_{i,j+1} - f_{i,j}| + \alpha^2\right\}. \tag{15}$$

A new estimate of $\mathbf{f}_{k+1}$ can be stated as a function of $\mathbf{f}_k$ with the help of this relationship:

$$\mathbf{f}^{k+1} = \left[\mathbf{H}^T\mathbf{H} + \lambda\mathbf{L}(\mathbf{f}^k)\right]^{-1}\mathbf{H}^T\mathbf{g}$$

$$\Rightarrow \mathbf{f}^{k+1} = \mathbf{f}^k - \left[\mathbf{H}^T\mathbf{H} + \lambda\,\mathbf{L}(\mathbf{f}^k)\right]^{-1}\nabla\mathbf{T}(\mathbf{f}^k) \quad . \tag{16}$$

The regularization operator $\mathbf{L}(\mathbf{f}^k)$ can be computed using the expression

$$L(\mathbf{f}) = D_x^T\,diag\left(\Psi'(\mathbf{f})\right)D_x + D_y^T\,diag\left(\Psi'(\mathbf{f})\right)D_y$$

$$= \begin{bmatrix}D_x^T & D_Y^T\end{bmatrix}\begin{bmatrix}diag\left(\Psi'(\mathbf{f})\right) & 0 \\ 0 & diag\left(\Psi'(\mathbf{f})\right)\end{bmatrix}\begin{bmatrix}D_x \\ D_y\end{bmatrix} \tag{17}$$

where

$$\Psi'_{i,j}(\mathbf{f}) = \Psi'\left(\left(D_{i,j}^x\mathbf{f}\right)^2 + \left(D_{i,j}^y\mathbf{f}\right)^2\right).$$

The desired value of $\mathbf{f}$ can be computed by means of several numerical algorithms. The Conjugate Gradient (CG) algorithm is the simplest one, and we will need the initial values of $\alpha$, $\mathbf{f_0}$, $\epsilon$, and $K$, where $\epsilon$ is the error tolerance between estimates and $K$ is the maximum number of iterations. $\mathbf{f_0}$ can be an even image (one color rectangle where all intensities have the value of the mean of the intensities). CG just computes the initial gradients and search for new direction to proceed. However, CG will converge slowly and linearly to the best solution. The above-mentioned procedure computes the conjugate gradient of (13) and gradually converges in a linear manner.

## 3. APPLICATIONS OF TV-NORM IN COMPUTER VISION
### 3.1 Computation Challenges
In order to minimize $J_{TV}(\mathbf{f})$, the gradient has to be computed. Differentiating $J_{TV}(\mathbf{f})$ with respect to $\mathbf{f}$ yields to the following nonlinear equation:

$$\mathbf{T}(\mathbf{f}) = \nabla J_{TV}(\mathbf{f}) = -\lambda\nabla\cdot\left(\frac{\nabla\mathbf{f}}{|\nabla\mathbf{f}|}\right) + \mathbf{H}^*\left(\mathbf{Hf} - \mathbf{g}\right) = \mathbf{0}. \tag{13}$$

The preceding minimization has the following computational challenges:

The operator $\nabla \cdot \left( \dfrac{\nabla \mathbf{f}}{|\nabla \mathbf{f}|} \right)$ is extremely nonlinear ; and

$\nabla \cdot \left( \dfrac{\nabla \mathbf{f}}{|\nabla \mathbf{f}|} \right)$ and $\mathbf{H}^* \mathbf{H}$ can be ill-conditioned which leads to numerical difficulties.

The conjugate gradient (CG) method can be used to solve (13). This procedure generates consecutive approximations of the estimation, the errors associated to the iterations, and the acceptable search directions used to revise all the required variables.

Although several schemes (Vogel & Oman, 1996) (Chambolle, 2004) have been devised to minimize $J_{TI}(\mathbf{f})$, it continues to be a time-consuming enterprise because it poses severe computational loads to problems with large $\mathbf{H}$ that lack some high-speed realization trick and/or suitable matrix representation.

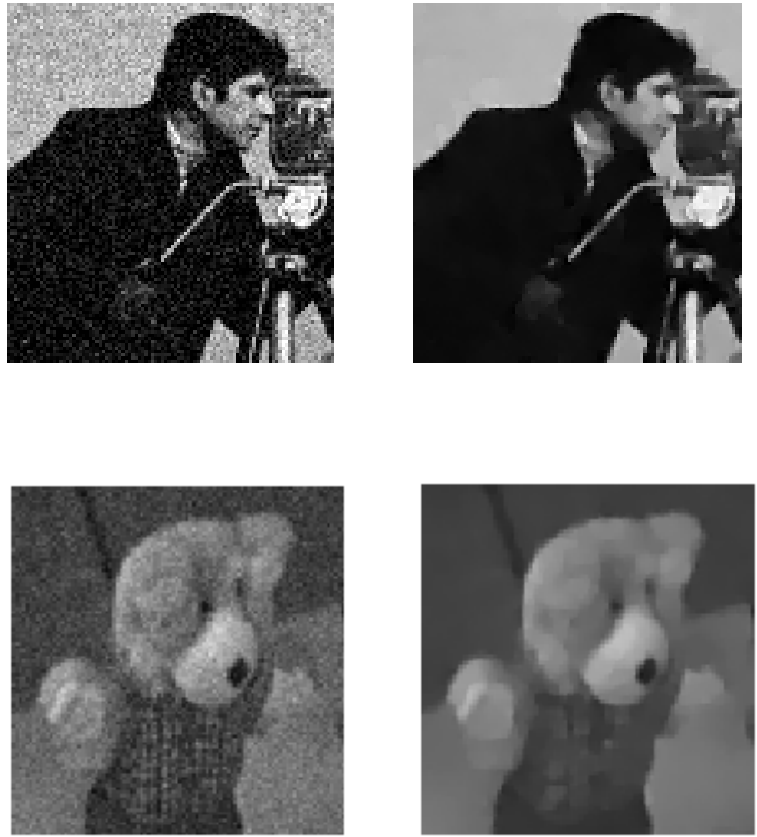

*Figure 1- The left column shows noisy images and the right side illustrate TV denoising by means of the TV algorithm proposed by (Chambolle, 2004)*

### 3.1.1 TV Denoising or Deconvolution

The application of Bayesian models to blind deconvolution is complicated when there is a difficult to handle probability density function (pdf) involving hidden variables **f** and **H** and the given observations **g** (Mateos, Molina, & Katsaggelos, 2005). This fact makes impossible the use simpler and less computationally demanding algorithms such as the EM technique. On the other hand, with variational approaches it is feasible to circumvent this problem.

The standard TV denoising problem has the form of (6) and it is a way of removing noise from images. It relies on the principle that signals with too much and perhaps spurious elements have high TV (the integral of the absolute gradient of the image has an elevated value). Hence, reducing the TV-norm of the image while keeping it very close to the original image, takes out unnecessary detail whereas conserving significant features such as boundaries (Rudin, Osher, & Fatemi, 1992).

This noise elimination procedure is better than simpler practices, which diminish noise though at the same time wipe out edges to a greater or less important extent. Moreover, TV-norm denoising is extremely successful at preserving boundaries while concomitantly eliminating noise in regions, regardless of the signal-to-noise ratio (SNR) (Strong & Chan, 2003) (Caselles, Chambolle, & Novaga, 2011).

Figure 1 shows an example of image denoising when white Gaussian noise is added to the original image with *a-priori* known noise.

### 3.1.2 TV Minimizing Blind Deconvolution

In the previous case, deconvolution was performed with the help of a known point spread function (PSF). Blind deconvolution (BD) allows the recuperation of an image from a defectively observed or unknown PSF of one or a set of several blurred images. BD estimates the PSF from an image or image set. Hence, there are no assumptions made about **H**. BD betters the PSF estimations and the true image **f** at each iteration. Popular BD methods comprise maximum a posteriori estimation (MAP) and expectation-maximization (EM) algorithms. A good initial PSF guess facilitates a faster convergence, although it is not indispensable.

(Money, 2006) broadened the minimizing functional proposed by (Chan & Wong, 1998) by adding a reference image to improve the quality of the estimated image and to reduce the computational load for the BD as follows:

$$T(\mathbf{f}) = \|\mathbf{Hf\text{-}g}\|^2 / 2 + \lambda_1 TV(\mathbf{f}) + \lambda_1 TV(\mathbf{f}). \tag{14}$$

Then, the problem can be recast as solving the equivalent Euler-Lagrange forms

$$1) \quad \mathbf{H}^* (\mathbf{Hf\text{-} g}) - \lambda_1 \nabla \cdot \left( \frac{\nabla \mathbf{f}}{|\nabla \mathbf{f}|} \right) = \mathbf{0} \quad , \text{ solve for } \mathbf{H}. \tag{15a}$$

$$2) \quad \mathbf{H}^* (\mathbf{Hf\text{-} g}) - \lambda_2 \nabla \cdot \left( \frac{\nabla \mathbf{f}}{|\nabla \mathbf{f}|} \right) = \mathbf{0}, \text{ solve for } \mathbf{f}. \tag{15b}$$

The subsequent algorithm was proposed by (Chan & Wong, 1998) and it is called the Alternating Minimization(AM) method.

*AM Algorithm*

1) Initial conditions: α, $\mathbf{f_0}$, ϵ, $\mathbf{H_0}$ and $K$, where ϵ is the error tolerance between estimates, $\mathbf{H}_0$ is the initial estimate of $\mathbf{H}$ and $K$ is the maximum number of iterations.

2) while (k < K) or ( ($\mathbf{f_k}$-$\mathbf{f_{k-1}}$)< ϵ do
      Solve  (15a) for $\mathbf{H_k}$.
      Solve (15b) for  $\mathbf{f_k}$.
  end

### 3.1.3 Image Restoration

The restoration obtained with regularization methods like RLS smoothes out the edges in the restored image. This can be alleviated by methods relying on $L_1$-norm regularization like TV regularization keeps the edges in the estimated image. The purpose is to recuperate a real image from an image distorted by several simultaneous phenomena such as blur and noise using the TV norm.

The image formation (forward process) is modeled mathematically using the expression

$$\mathbf{g=Hf+n},$$

where, $\mathbf{g}$ is the observed distorted and noisy image,  $\mathbf{H}$ represents  some Point Spread Function (PSF) or blurring function, $\mathbf{f}$ is the original  image, $\mathbf{n}$ is white Gaussian noise.

The Least Absolute Shrinkage and Selection Operator (LASSO) estimator is a shrinkage and selection procedure relying on the $L_1$ norm for linear regression created by (Tibshirani, 1996). The LASSO estimate is calculated by means of the minimization of a quadratic problem consisting of the customary sum of squared errors, with a bound on the summation of the absolute values of the coefficients $f_i$ and it is defined by

$$\sum_i \left( g_i - \sum_j H_{ij} f_j \right)^2 \text{ subjected to } \sum_j | f_j | \le c ,$$

where $c$ is a parameter controlling the amount of regularization.

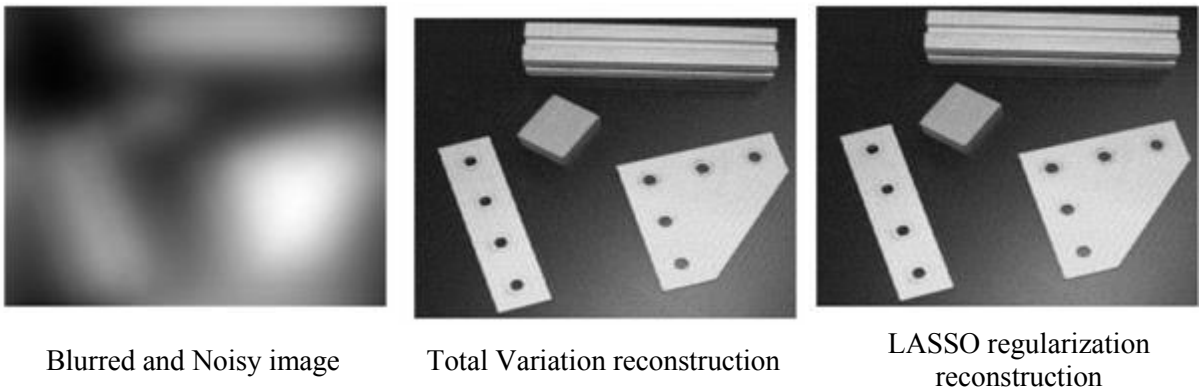

Blurred and Noisy image     Total Variation reconstruction     LASSO regularization reconstruction

*Figure 2. Reconstruction using Total Variation regularization and LASSO regularization*

TV techniques conserve edge information in computer vision algorithms at the expense of a high computational load. Restoration time in $L_1$-based TV regularization, for instance, is higher than when LASSO is used. Restoration time using and image quality are respectively lower and better with LASSO than with $L_1$-based TV regularization.  Hence, for some settings, LASSO provides an noteworthy alternative to the  $L_1$-TV norm.  Studies illustrate that an augment in the amount of blur amplifies the restoration error when the $L_1$ is employed. An increase in the noise level exerts a significant influence on the residual error (Agarwal, Gribok, & Abidi, 2007). Nevertheless, since there are other ways of calculating the TV norm, the computation time and the estimation quality can be further improved and outdo LASSO. Figure 2, shows restored images obtained from an observed one subjected to blurring and noise.

### 3.1.4 Optical Flow Estimation

In computer vision, the existent motion estimation problem in a video sequence was first studied by (Horn & Schunck, 1981) and it can bring in lots of information to help understanding a given system, scenario and/or problem. Characteristically, the goal is to identify the displacement vector field (DVF) involving successive frames also known as optical flow (OF). On the other hand, variational techniques are very important, and allow for accurate estimation of the DVFs while rooted in the minimization of functionals.

Consider an image sequence f($x$, $y$, $t$), where ($x$, $y$) stands for the location within a frame domain $\Omega$, and $t \in [0, T]$ indicates time. After that, the postulation of unchanging brightness along time (invariance of the displaced frame difference, also known as DFD) can be stated as

$DFD$(f, $u$,$v$) = f($x + u$, $y + v$, $t + 1$)    f($x$, $y$, $t$)=$0$.           (16)

With the help of a Taylor expansion and after dropping all higher order terms, one obtains its linearized form, the so-called optic flow constraint (OFC)

$f_x\, u$ +$f_y\, v$ + $f_t$ =$0$.           (17)

Here, the function $\mathbf{u}$($x$, $y$, $t$)= ($u$ ,$v$) is the wanted DVF and subscripts denote partial derivatives. There are different types of regularization for the non-unique solution of the OF problem:

   i) Uniform regularization takes for granted an overall smoothness constraint and it does not adjust itself to semantically noteworthy image and/or OF arrangements;
   ii) image-driven regularization that assumes piecewise smoothness and respects discontinuities in the image (Nagel & Enkelmann, 1986); and
   iii) OF regularization assumes piecewise smoothness and respects borders in the DVF as in (Cohen, 1993) (Weickert & Schnörr, 2001).

Variational methods are among the finest techniques for estimating the OF by means of error evaluation procedures, however they are frequently slow for real-time applications. For the most part, the computational costs for solving the nonlinear system of equations via typical numerical methods are considered significantly elevated. Variational schemes relying on bidirectional multigrids generate a refined hierarchy of equation systems with first-rate error decrease.

OF algorithms based on variational approaches have been gaining lots of popularity, because they handle dense flow fields and their performance can be good if spatial and temporal discontinuities are retained in the video. Regrettably, this flexibility implies elevated computational load, but this does not preclude real-time performance if methods such as multigrid are employed (Bruhn, Weickert, Kohlberger, & Schnörr, 2005).

Variational computer vision algorithms belong to one of the following classes: i) anisotropic image-driven techniques as proposed by (Nagel & Enkelmann, 1986) which results in a linear system of equations; and ii) isotropic OF-driven schemes with TV regularization that involve solving a nonlinear system of equations.

$$J(u,v) = \int_{\Omega} ((f_x u + f_y v + f_t)^2 + \lambda (\nabla u^T D(\nabla f)\nabla u + \nabla v^T D(\nabla f)\nabla v))dxdy \,, \tag{18}$$

where $\nabla = (\delta x,\ \delta y)^T$ denotes the spatial gradient and $\mathbf{D}(\nabla \mathbf{f})$ is a projection matrix perpendicular to $\nabla \mathbf{f}$ that is defined as

$$D(\nabla f) = \frac{1}{|\nabla f| + 2\varepsilon^2} \begin{bmatrix} f_y^2 + \varepsilon^2 & -f_x f_y \\ -f_x f_y & f_x^2 + \varepsilon^2 \end{bmatrix} = \begin{pmatrix} a & b \\ b & c \end{pmatrix}. \tag{19}$$

$\varepsilon$ serves as a parameter that prevents matrix $\mathbf{D}(\nabla \mathbf{f})$ from being singular. The minimization of this convex functional comes down to solving the following equations

$$f_x^2 u + f_x f_y v + f_x f_t - \frac{\lambda}{2} L_{AN}(u,v) = 0 \,, \text{ and} \tag{20}$$

$$f_x f_y u + f_y^2 v + f_y f_t - \frac{\lambda}{2} L_{AN}(v,u) = 0 \,, \tag{21}$$

with $\quad L_{AN}(z(x,y),\tilde{z}(x,y)) = div(D(\nabla z(x,y), \nabla \tilde{z}(x,y))\nabla z(x,y))\,. \tag{22}$

In contrast to image-driven regularization methods, OF-driven techniques trim down smoothing where edges in the flow field occur during computation. (Drulea & Nedevschi, 2011) proposed for this class of variational OF techniques an isotropic method that penalizes deviations from the smoothness constrain with the $L_1$-norm of the flow gradient magnitude. This strategy matches TV regularization and it can be linked to norms that are statistically robust to error. In that way, large variations are penalized more mildly than what happens when the popular $L_2$-norm is used. Therefore, regions with large gradients as it is the case with edges are better handled. Rewriting (18) yields

$$J(u,v) = \int_{\Omega} ((f_x u + f_y v + f_t)^2 + \lambda \sqrt{|\nabla u|^2 + |\nabla v|^2 + \varepsilon^2})dxdy \,,$$

where $\varepsilon$ serves as small control parameter to avoid having a zero denominator in (19). Another functional that also provides a TV regularization estimate is proposed in (Drulea & Nedevschi, 2011). Apparently, the consequent Euler-Lagrange equations given by

$$f_x^2 u + f_x f_y v + f_x f_t - \frac{\lambda}{2} L_{TV}(u,v) = 0 \,, \text{ and}$$

$$f_x f_y u + f_y^2 v + f_y f_t - \frac{\lambda}{2} L_{TV}(v,u) = 0$$

are very similar in structure to (20)-(21). However,

$$L_{TV}(z(x,y),\tilde{z}(x,y)) = div(D(\nabla z(x,y), \nabla \tilde{z}(x,y))\nabla z(x,y))$$

is clearly a nonlinear differential operator in $z$ and $\tilde{z}$ , since

$$D(\nabla z, \nabla \tilde{z}) = \frac{1}{\sqrt{|\nabla z|^2 + |\nabla \tilde{z}|^2 + \varepsilon^2}} I \, ,$$

with $\mathbf{I}$ is the identity matrix, $b=0$ and $c=a$ . Soon, it will become clear that the differential operator $L_{TV}$ is nonlinear and that it impacts seriously the resultant discrete system of equations.

A suitable discretization for the previous Euler-Lagrange equations can be obtained via unknown functions $u(x, y, t)$ and $v(x, y, t)$ on a grid with pixel size $\mathbf{h}=(h_x, h_y)^T$ , where $u_{i,j}{}^h$ stands for the approximation to $\mathbf{u}$ at some pixel located at $(i,j)$ with $i=1, \ldots, N_x$ and $j=1, \ldots, N_y$ . Spatial and temporal derivatives of the image data and discretized versions of the operators $L_{AN}$ and $L_{TV}$ are approximated using finite differences as follows:

$$f_{xi,j}^{2,h} u_{i,j}^h + f_{xi,j}^h f_{yi,j}^h v_{i,j}^h + f_{xi,j}^h f_{ti,j}^h - \lambda L_{ANi,j}^h u_{i,j}^h = 0$$

$$f_{(xi,j)}^h f_{(yi,j)}^h u_{(i,j)}^h + f_{(yi,j)}^{(2,h)} v_{(i,j)}^h + f_{(yi,j)}^h f_{(ti,j)}^h - \lambda L_{ANi,j}{}^h v_{i,j}^h = 0$$

where the operator $L_{ANij}{}^h$ indicates $L_{AN}$ discretized at some pixel located at $(i,j)$. The previous expressions amount to a linear system of $2N_xN_y$ equations in $u_{i,j}{}^h$ and $v_{i,j}{}^h$ . Discretizing the Euler-Lagrange equations for the corresponding TV-based method leads to the nonlinear system of equations shown underneath

$$f_{xi,j}^{2,h} u_{i,j}^h + f_{xi,j}^h f_{yi,j}^h v_{i,j}^h + f_{xi,j}^h f_{ti,j}^h - \lambda L_{TVi,j}^h u_{i,j}^h = 0$$

$$f_{(xi,j)}^h f_{(yi,j)}^h u_{(i,j)}^h + f_{(yi,j)}^{(2,h)} v_{(i,j)}^h + f_{(yi,j)}^h f_{(ti,j)}^h - \lambda L_{TVi,j}^h v_{i,j}^h = 0$$

Here the finite difference approximation of $L_{TV}(u, v)$ and $L_{TV}(v, u)$ yields the product of a common nonlinear operator $L_{TVi,j}{}^h (u_{i,j}{}^h, v_{i,j}{}^h)$ and the pixels $u_{i,j}{}^h$ and $v_{i,j}{}^h$ , respectively.

### 3.2 Solutions and Recommendations

TV norms can be discretized differently from what was shown in previous sections if finite differences, with atypical geometric arrangements of close pixels involving 3, 4 or 8 neighbors, and/or special norms are used. In Section 2, the TV norm regularization was stated so that it could benefit from the fast algorithm proposed by (Chambolle, 2004).

A discretization procedure is an approximated representation of real continuous signals. Since the pixel dimension cannot usually be selected in computer vision, presupposing it is small enough may not be appropriate. TV-norms founded on finite differences are also arguable due to the need of having sub-pixel accuracy in algorithms that require, for example, sub-pixel interpolation. According to the sampling theory, the discretization procedure shown in Section 2 is not good, because notwithstanding the fact that $\mathbf{f}$ was sampled consistently with Shannon's theorem, the squares in $|\nabla \mathbf{f}|$ bring in high frequencies that require smaller sampling intervals to be attenuated. Therefore, the estimation of $|\nabla \mathbf{f}|$ has problems due to alias and the resulting TV norm estimate of $\mathbf{f}$ will carry artifacts.

Despite the fact that using TV-norm can decrease fluctuations and improve regularization of inverse problems in computer vision without compromising edges, it has some undesirable side effects:

i. The solution to the (Rudin, Osher, & Fatemi, 1992) model is prone to contrast loss due to scaling of the regularization and fidelity terms because it decreases the bounded TV norm of a function, in the vicinity of its mean. In general, a reduction of the contrast decreases the regularization term of the (Rudin, Osher, & Fatemi, 1992) model and boosts the fidelity term.

ii. Geometric alterations may perhaps appear since the TV norm of a function is reduced once the length of all level sets is decreased. Sometimes, this distorts silhouettes that are not part of the shape-invariant set when the (Rudin, Osher, & Fatemi, 1992) model is employed. Still, for circular parts, (Strong & Chan, 2003) have demonstrated that shape is kept for a small variation in λ as well as location albeit in the presence of moderate noise. Corners may suffer deformation as well.

iii. Staircasing refers to the case when the estimated image may appear blocky outside corners due to the high values of the level sets curvature. TV norm amendments, which include higher-order derivatives, are an alternative to this problem when there is sensible parameter selection.

iv. Even though extremely valuable, the TV norm cannot always keep textures because the model from (Rudin, Osher, & Fatemi, 1992) has the propensity to affect small features present in images and it can suffer because of scaling. Hence, the net effect is texture loss.

The NUMIPAD library has a collection of techniques to solve inverse problems such as Tikhonov regularization, Total Variation, Basis Pursuit, etc. (Rodrıguez & Wohlberg). Other very good package written in MATLAB is the $L_1$-magic (Candes, Romberg, & Tao).

## FUTURE RESEARCH DIRECTIONS

For models such as $\mathbf{g} = \mathbf{Hf} + \mathbf{n}$, it is not viable to state unambiguously the probabilistic relationship associated to the convolving functions when $\mathbf{g}$ is known and Bayesian inference is used (Likas & Galatsanos, 2004) (Mateos, Molina, & Katsaggelos, 2005). A variational scheme can help to overcome this hindrance with higher performance than conventional techniques. The principal deficiency of the variational line of reasoning is the lack of systematic evaluation procedures to appraise the variational bound stiffness. Evidently, more studies on this topic and optimization procedures are required. Still, the suggestd method is rather extensive, so that it can be combined with other Bayesian models for several imaging applications.

It is a well-known fact that LS estimate is not robust to outliers. There are some recent efforts in compressed sensing (Candes. E.J. & Wakin, 2008) (Candes, Romberg, & Tao, 2006) that focus on the $L_0$ -$L_1$ equivalence to determine the gradient which best replaces the degraded gradient field. These works investigate robust strategies to estimate gradient by taking into consideration error corrections along with concepts from research on sparse signal retrieval. Among other things, they confirm that the location of errors is as important as the number of errors when it comes to the gradient integration required by the TV norm.

To reconcile TV with Shannon Theory (Shannon, 1948), the TV of a discrete image $\mathbf{f}$ can be defined as being the exact (continuous) TV of its Shannon interpolate $\mathbf{F}$ which is equal to the Fourier transform of $\mathbf{f}$.

However, since TV(**F**) cannot be computed exactly, (Moisan, 2007) uses a Riemann sum with an oversampling factor $n$, and define the Spectral Total Variation (STV) of **f** (of order $n \geq 1$) in this manner:

$$STV_n(\mathbf{f}) = (1/n^2) \Sigma_{0 \leq k, l < nN} [|DU| \, (k/n, l/n)].$$

$STV_n(\mathbf{f})$ is supposed to yield a fine estimate of $TV(\mathbf{F})$ for any **f**, given that this measure is a regularization term. $n=1$ is not a good option, because controlling the gradient norm of **F** only at grid points does not permit its control between grid points. When $n=2$, a new TV discretization is obtained along with several improvements: grid independence, compatibility with Shannon theory and the possibility of achieving sub-pixel precision, at the expense of applying Fourier Transforms, which is a widespread necessity in deconvolution problems, for instance.

Adaptive TV norm calculations are needed in order to better care for texture along with fine-scale details. Afterwards, the adaptive procedure enforces local constraints depending on local metrics (Gilboa, Sochen, & Zeevi, 2003).

Despite the fact that most works concentrate on scalar functions, the extension of the reasoning used in this chapter to color or multi-channel images remains an important challenge because it requires vector valued parameters and/or functions. This generalization is not trouble-free, but it can result from geometric measure theory.

The improvement of proper multigrid approaches becomes more complicated thanks to the anisotropy and/or nonlinearity of the basic regularization strategies, but they can lead to real-time performance.

## CONCLUSION

TV regularization is a broadly applied method because it keeps image edges. Developments on this technique are centered mostly on the application of higher order derivatives (Chan, T.; Esedoglu, S.; Park, F.; Yip, 2005) (Stefan, W.; Renaut, R.; Gelb, 2010) (A Chambolle & Lions, 1997) (Yuan, Schnörr, & Steidl, 2009) (Chan, Marquina, & Mulet, 2000) and on nonlocal simplifications too. The fundamental idea behind TV regularization can benefit from the use of a more general differential operator. This increases flexibility because it accounts for the occurrence of a linear system and assorted inputs.

In view of the fact that there exist further ways of estimating the divergence operator required by the TV norm, the totaling time and image quality can be made superior.

To augment the efficiency of the regularization procedure, alternative algorithms relying on dual forms of the TV norm call for further research on topics such as exponential spline wavelets (Khalidov & Unser, 2006) or generalized Daubechies wavelets (Schwamberger, Le, Schölkopf, & Franz, 2010) (Vonesch, Blu, & Unser, 2007) (Subashini & Krishnaveni, 2011b). Undeniably, these wavelet improvements can be tuned to a given differential operator and their use for regularized computer vision purposes corresponds to a synthesis prior (Franz & Schölkopf, 2006). Prospective research is also wanted to strengthen a more precise relationship involving discrete domain approaches and suitable forms of TV regularization in the continuous domain. Current work (Vonesch, Blu, & Unser, 2007) has shown that the usual TV norm calculation via the $L_1$-norm with finite differences can be associated to appropriate representations of stochastic processes.

Compressed sensing (CS) enables recovery of compressed images from a small number of linear measurements (Kienzle, Bakir, Franz, & Schölkopf, 2005) (Candes. E.J. & Wakin, 2008). It can be an alternative to methods that try to handle missing information, but I involve larger image representations such as (Subashini & Krishnaveni, 2011). It had been well known that without noise contamination, images with completely sparse gradients can be recovered with a high degree of accuracy through TV-norm. Hence, there are several CS algorithms relying on TV regularization because according to (Candes, Romberg, & Tao, 2006) they have outstanding outputs in the presence of images with sparse discrete gradients. TV methods also work well with piecewise constant images. Furthermore, since images are easier to compress when the discrete gradient representation is used, the TV-norm has advantages over wavelets in the presence of additive and/or quantization noise (Jiang, Li, Haimi-Cohen, Wilford, & Zhang, 2012).

Noise can obliterate image analysis. Images with unknown noise can be handled with no priors if a wavelet decomposition technique is used with a non-isotropic TV filtering in a way that there is gain from both the multiresolution capacity of the wavelet transform (Subashini & Krishnaveni, 2011b) as well as the edge-preserving properties of the TV-norm (Subashini, Krishnaveni, & Thakur, 2010) (Zhang, 2009).

## ADDITIONAL READING SECTION